\documentclass[sn-mathphys,Numbered]{sn-jnl}

\usepackage{graphicx}%
\usepackage{multirow}%
\usepackage{amsmath,amssymb,amsfonts}%
\usepackage{amsthm}%

\usepackage[title]{appendix}%
\usepackage{xcolor}%
\usepackage{textcomp}%
\usepackage{manyfoot}%
\usepackage{url}%
\usepackage{algorithm}%
\usepackage{algorithmicx}%
\usepackage{algpseudocode}%
\usepackage{listings}%
\usepackage{bm}
\usepackage{algpseudocode}
\usepackage{lipsum}
\usepackage{lineno,dsfont}
\usepackage{amscd}
\usepackage{stmaryrd}
\usepackage{ifthen}
\usepackage{subcaption}
\usepackage{epsfig,wrapfig,epic}
\usepackage{dsfont}
\usepackage[none]{hyphenat}
\usepackage{hyperref}

\newtheorem{example}{Example}[section]

\newtheorem{definition}{Definition}[section]
\newtheorem{theorem}{Theorem}[section]

\usepackage{lmodern}



\begin{document}

\title[Double Coupling Architecture and Training Method for Optimization Problems of Differential Algebraic Equations
with Parameters]{Double Coupling Architecture and Training Method for Optimization Problems of Differential Algebraic Equations
with Parameters}


\author[1,2]{\fnm{Wenqiang} \sur{Yang}}\email{yangwenqiang@cigit.ac.cn}

\author*[1]{\fnm{Wenyuan} \sur{Wu}}\email{wuwenyuan@cigit.ac.cn}

\author[1]{\fnm{Yong} \sur{Feng}}\email{yongfeng@cigit.ac.cn}

\author[1]{\fnm{Changbo} \sur{Chen}}\email{chenchangbo@cigit.ac.cn}

\affil*[1]{\orgdiv{Chongqing Key Laboratory of Automated Reasoning and Cognition}, \orgname{Chongqing Institute of Green and Intelligent Technology, Chinese Academy of Sciences}, \orgaddress{\street{266 Fangzheng Avenue}, \city{Liangjiang District}, \postcode{400714}, \state{Chongqing}, \country{China}}}

\affil[2]{\orgdiv{Chongqing School}, \orgname{University of Chinese Academy of Sciences}, \orgaddress{\street{266 Fangzheng Avenue}, \city{Liangjiang District}, \postcode{400714}, \state{Chongqing}, \country{China}}}

\begingroup
\renewcommand\thefootnote{}
\footnotetext{The research is supported by the National Key Research Project of China under Grant No.2023YFA1009402.}
\endgroup

\abstract{
Simulation and modeling are essential in product development, integrated into the design and manufacturing process to enhance efficiency and quality. They are typically represented as complex nonlinear differential algebraic equations. The growing diversity of product requirements demands multi-task optimization, a key challenge in simulation modeling research.
A dual physics-informed neural network architecture has been proposed to decouple constraints and objective functions in parametric differential algebraic equation optimization problems. Theoretical analysis shows that introducing a relaxation variable with a global error bound ensures solution equivalence between the network and optimization problem. A genetic algorithm-enhanced training framework for physics-informed neural networks improves training precision and efficiency, avoiding redundant solving of differential algebraic equations. This approach enables generalization for multi-task objectives with a single, training maintaining real-time responsiveness to product requirements.
}


\maketitle
\section{Introduction}\label{sec:intro}

Simulation and modeling serve as core technologies in product development, deeply integrated into the entire design and manufacturing process, and are pivotal in enhancing development efficiency and quality. Simulation models can generally be abstracted as differential algebraic equations (DAE)~\cite{Brenan95}, characterized by complex nonlinear coupling relationships among system variables. Furthermore, the growing diversification of product requirements demands systematic approaches to handle multi-task optimization during product design processes. This scenario, formally defined in Definition~\ref{def:opt_dae}, involves systems with fixed constraints but dynamically changing objective functions.

\begin{definition}\label{def:opt_dae}
The optimization problem for parametric differential-algebraic equations can be described as:
\begin{equation}\label{opt_goal}
\begin{array}{cc}
   \text{obj} & \underset{\bm{p}}{{\min }}\,J\left( \bm{p} \right)=\int_{{{t}_{0}}}^{{{t}_{f}}}{\bm{\lambda }\left( s,\bm{p}, \bm{x}\left( s,\bm{p} \right) \right)}ds + \bm{\varphi }\left( \bm{p},\bm{x}\left( {{t}_{f}},\bm{p} \right) \right)  \\
   \text{s.t.} & \left\{ \begin{matrix}
                     \bm{\Phi}\left( t,\bm{p},\bm{x}\left( t,\bm{p} \right),\cdots,\bm{x}^{(\bm{\ell})}\left( t,\bm{p} \right) \right)=\bm{0}, \\
                      \bm{x}\left( {{t}_{0}},\bm{p} \right)={{\bm{x}}_{0}}\left( \bm{p} \right), \\
                      t\in \left[ {{t}_{0}},{{t}_{f}} \right],\ \bm{p}\in \left[ {{\bm{p}}^{L}},{{\bm{p}}^{U}} \right].
                     \end{matrix} \right.
  \end{array}
\end{equation}
\end{definition}

Here, $t \in \mathbb{I} \subseteq \mathbb{R}$ denotes the independent variable, and $\bm{p} \in P \subset \mathbb{R}^{m}$ represents the parameter vector with lower and upper bounds $\bm{p}^{L}$ and $\bm{p}^{U}$, respectively. The state variable $\bm{x}(t,\bm{p}) = (x_1,\dots,x_n) \in \mathbb{R}^n$ denotes the system response, and $\bm{x}^{(\bm{\ell})}$ represents derivatives up to order $\bm{\ell}=(\ell_1,\dots,\ell_n)$. The function $\bm{\Phi}$ defines the DAE constraints, which may include both differential and algebraic components. The running cost $\bm{\lambda}(t,\bm{p},\bm{x})$ and terminal cost $\bm{\varphi}(\bm{p},\bm{x}(t_f,\bm{p}))$ together form the objective functional $J(\bm{p})$.

With multi-task requirements, such optimization problems significantly increase computational complexity and impose higher demands on both generality and efficiency, becoming a critical challenge in simulation-driven design.

\subsection{Optimization and Learning Methods for Parametric DAEs}

Traditional approaches for parametric DAE optimization include indirect methods such as variational techniques~\cite{Betts2010}, as well as direct transcription-based strategies including simultaneous methods~\cite{BIEGLER20071043} and finite difference schemes~\cite{MAZUMDER201651}. These methods typically compute a single optimal solution for a given objective and rely on first-order optimality conditions. Global optimization methods extend this framework by seeking globally optimal solutions. Deterministic approaches, such as branch-and-bound and interval analysis~\cite{Scott2015}, provide theoretical guarantees by exploiting structural properties of the system~\cite{Papamichail2002ARG,XUJuan2025}, but their computational complexity grows rapidly with system dimension and nonlinearity, which restricts their applicability to low-dimensional or weakly nonlinear systems. Stochastic methods, including simulated annealing~\cite{Kirkpatrick1983,LI2021203}, particle swarm optimization~\cite{Abed2021}, and ant colony optimization~\cite{ZHOU2022105139,Zhang2021}, offer greater flexibility for nonlinear DAEs but suffer from high computational cost and lack of convergence guarantees.

To improve computational efficiency, neural network-based approaches have been introduced as surrogate models that enable offline learning and rapid online evaluation. Early work by Lagaris et al.~\cite{Lagaris98} employed feedforward neural networks to approximate solutions of differential equations via residual minimization. Neural ODEs~\cite{chen2018} further enhanced modeling capability by integrating neural networks with numerical solvers, but their applicability remains largely restricted to ODE systems and is affected by error accumulation during time integration.

Physics-informed neural networks (PINNs)~\cite{RAISSI2019686} provide a unified framework by embedding governing equations directly into the loss function, allowing simultaneous data fitting and constraint enforcement. For DAE systems, the DAE-PINN framework~\cite{moya2023} extends PINNs by decoupling differential and algebraic components using parallel subnetworks, enabling the solution of low-index DAE systems. Subsequent developments improve performance on high-index and stiff systems through architectural and training enhancements. However, most existing PINN-based approaches rely on soft penalty formulations, which approximate constraints through weighted residuals and may lead to constraint violations in strongly coupled DAE systems.

To address this limitation, hard-constrained formulations have been proposed to enforce constraints more rigorously. For instance, hPINNs~\cite{Lu2021hPINN} incorporate equality constraints directly into the solution space, while recent work such as DAE-HardNet~\cite{Golder2025DAEHardNet} introduces projection-based mechanisms that explicitly enforce algebraic constraints during training. These approaches significantly improve the feasibility and stability of learned solutions, especially for systems with strong algebraic coupling.

In parallel, PINNs have been extended to solve optimization problems constrained by differential equations. Control-oriented PINN frameworks incorporate optimality conditions into the training process, enabling simultaneous learning of state and control variables. Bi-level PINNs~\cite{BilevelPINN2022} separate optimization and constraint satisfaction into hierarchical structures, allowing efficient gradient propagation, while ADMM-based PINNs~\cite{Song2023ADMM_PINN} decompose optimization problems into subproblems to enhance scalability. Nevertheless, these approaches are primarily developed for PDE-constrained optimization and lack systematic treatment for parametric DAE systems, particularly under multi-task settings.

Moreover, the inherent nonlinearity of parametric DAEs may lead to degeneration caused by singular Jacobian matrices, resulting in the loss of physical constraint information~\cite{Yang2024}. Existing methods lack effective mechanisms to detect and regularize such degradation. At the same time, most approaches focus on solving a single optimization task and require repeated retraining when objective functions change, leading to significant computational overhead. These limitations highlight the need for new frameworks that can simultaneously ensure constraint satisfaction, preserve physical information, and enable efficient multi-task optimization.

\subsection{Contributions}

To address the above challenges, this paper proposes a novel learning-based framework for parametric DAE optimization. The main contributions are summarized as follows.

First, a dual-PINN decoupling strategy is developed, where one network learns the constraint manifold of the DAE system and its representation is embedded into another network that solves optimization problems with varying objective functions. This mechanism eliminates repeated DAE solving and enables efficient multi-task optimization.

Second, an embedding-based mechanism~\cite{Yang2024} is introduced to detect and regularize parametric degradation, which improves the quality of training data and preserves essential physical information.

Third, an equivalence guarantee is established through the introduction of relaxation variables, and a tight global error bound~\cite{Yang2021} is derived to ensure that the proposed composite network shares the same solution set as the original optimization problem.

Finally, a hybrid training strategy combining global exploration via genetic algorithms and local refinement is proposed to balance computational efficiency and solution accuracy.

\section{Preliminaries}\label{sec:pre}

\subsection{PINNs for Optimization}

\begin{figure}[!h]
\centering
\includegraphics[width=\textwidth]{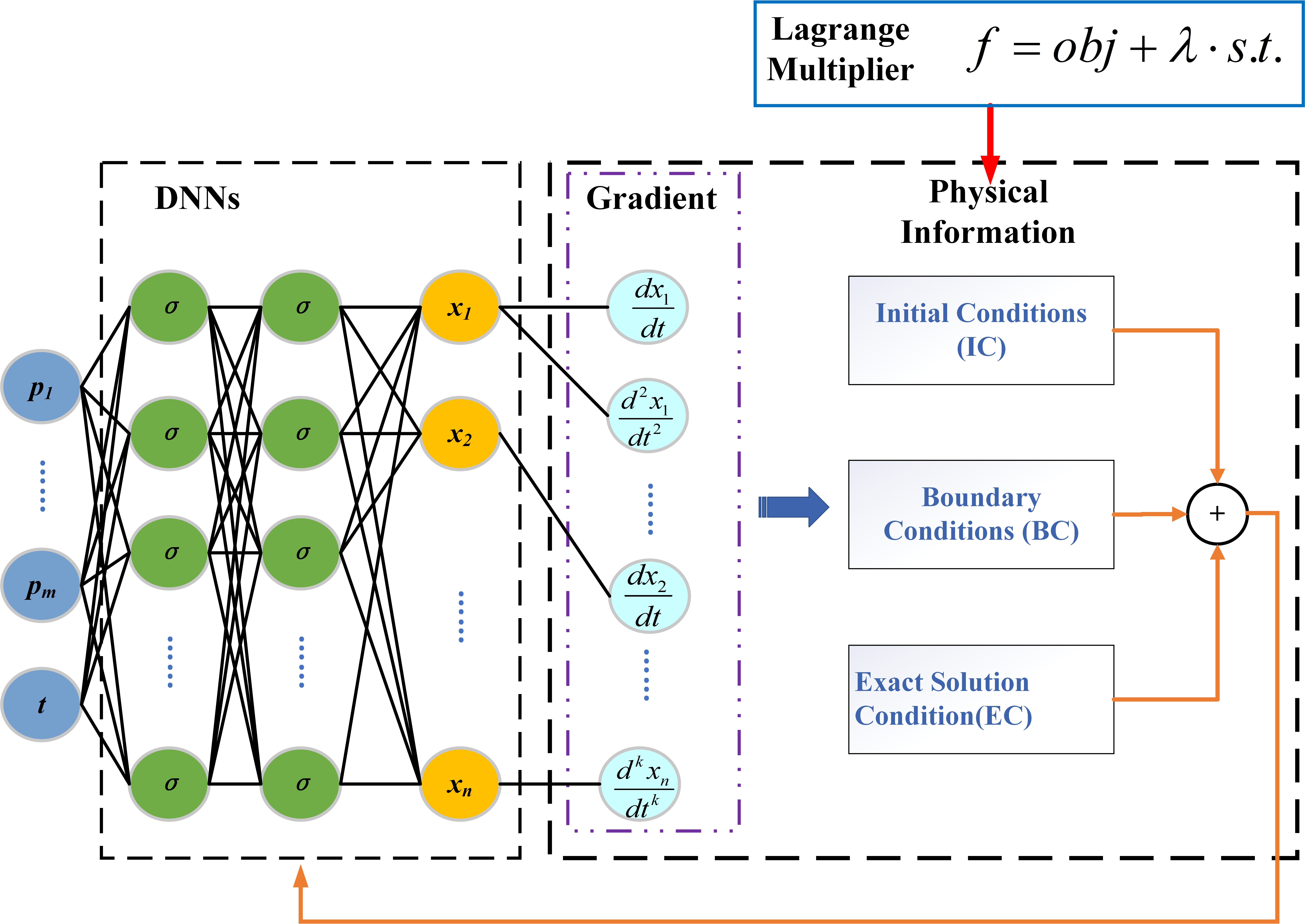}
\caption{PINNs for Optimization}\label{fig:pinns}
\end{figure}

For parametric ordinary differential equation (ODE) constrained optimization problems, the formulation is given by
\begin{equation*}
    \min_{\bm{p}} J(\bm{p}) \quad \text{s.t.} \quad \frac{d\bm{x}(t,\bm{p})}{dt} = f(\bm{x}(t,\bm{p}), t; \bm{p}), \quad \bm{x}(t_{0},\bm{p})=\bm{x}_{0}(\bm{p}).
\end{equation*}

Physics-informed neural networks (PINNs) incorporate both the governing equations and objective function into a unified loss function, where the network parameters and optimization variables are updated simultaneously via gradient-based methods~\cite{RAISSI2019686}. The neural network acts as a surrogate model for the system dynamics, enabling direct optimization without repeatedly solving the differential equations.

For parametric DAE systems, the constraints in Equation~(\ref{opt_goal}) can be approximated similarly by introducing a neural network
\begin{equation}
\overline{\bm{x}} = \mathcal{NN}(t, \bm{p}; \bm{\theta}),
\end{equation}
where $\overline{\bm{x}}$ approximates the solution $\bm{x}(t,\bm{p})$. The training process typically involves three types of data: initial condition samples enforcing $\bm{x}(t_0,\bm{p})$, collocation points minimizing the DAE residuals via automatic differentiation, and supervised solution data when available. The overall loss is constructed as a weighted combination of these components.

However, for parametric DAE systems, accurate supervised data is often unavailable, especially under parameter variations. In the presence of algebraic constraints, multiple solution branches may exist, making the learning problem ill-posed. As a result, PINNs may converge to incorrect local minima when trained in an unsupervised manner.

\begin{example}\label{ex:song}
Consider a cantilever beam with a parameter $p$ affecting the deflection $y(x,p)$. The optimization problem is
\[
\begin{matrix}
  obj & \underset{\bm{p}}{{\min }}\,J(\bm{p}) = \min\limits_{p}{\int_{0}^{5}{(p-3.1)(p-3.3)(p-3.6)(p-3.8)\cdot y_{2}(x,p)}dx} \\
  s.t. & \left\{\begin{array}{c}
{\frac {\rm d^{2}}{{\rm d}x^{2}}}(y_{1}+y_{2}) + \frac{1}{p+1}(1-\sin(x))+y_{1} = 0\\
 y_{1}^{2} - y_{2}^{2} = 0\\
 y_{1}(0,p) = -\frac{1}{p+1},\quad
 y_{2}(0,p) = \frac{1}{p+1}\\
 x \in [0, 5],\ p \in [3, 4] 
\end{array}\right. 
\end{matrix}
\]
\end{example}

\begin{figure}[!ht]
    \centering
    \begin{subfigure}[t]{0.48\linewidth}
        \centering
        \includegraphics[width=0.93\linewidth]{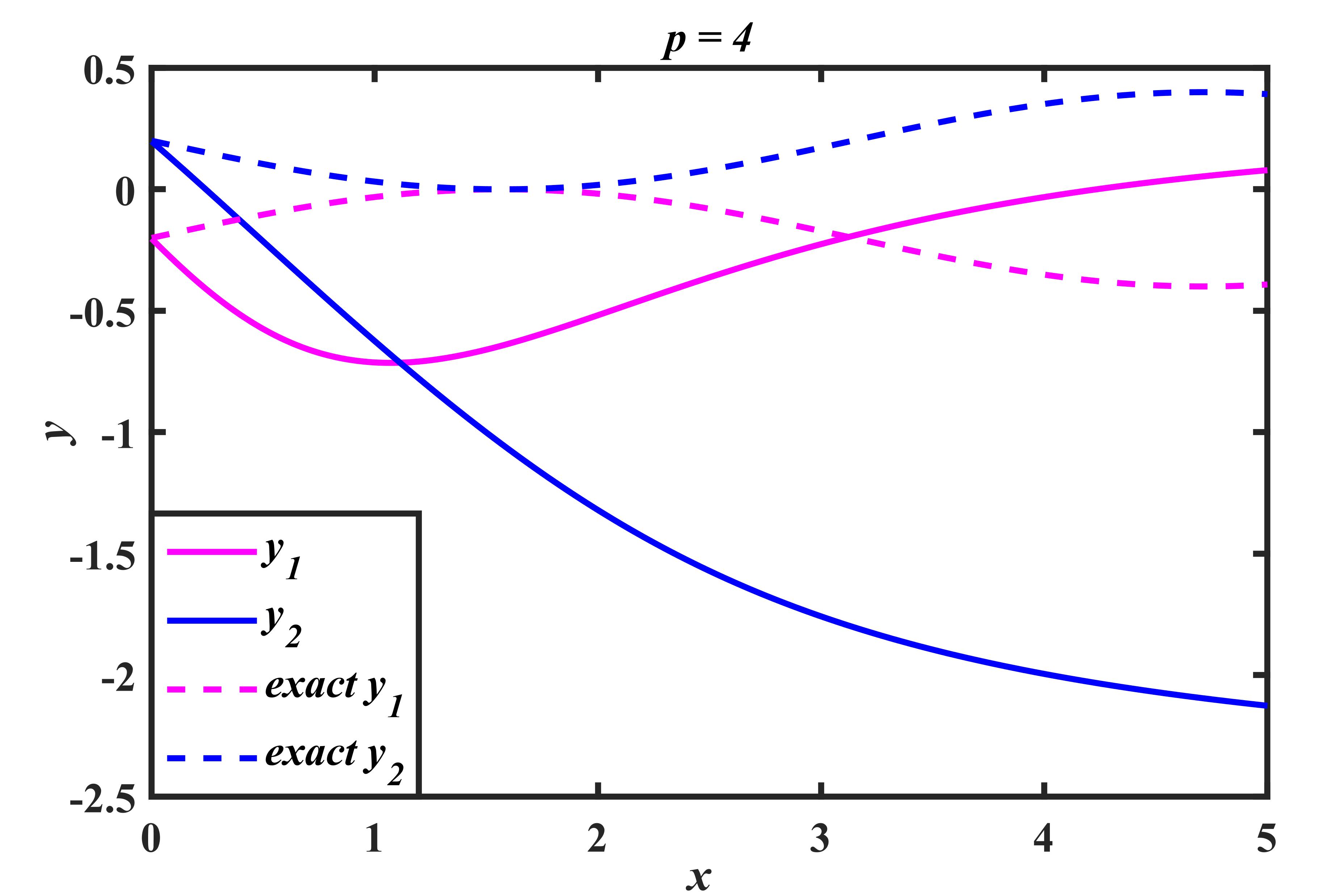}
        \caption{Unsupervised}
        \label{fig:ex_origin}
    \end{subfigure}
    \hfill
    \begin{subfigure}[t]{0.48\linewidth}
        \centering
        \includegraphics[width=\linewidth]{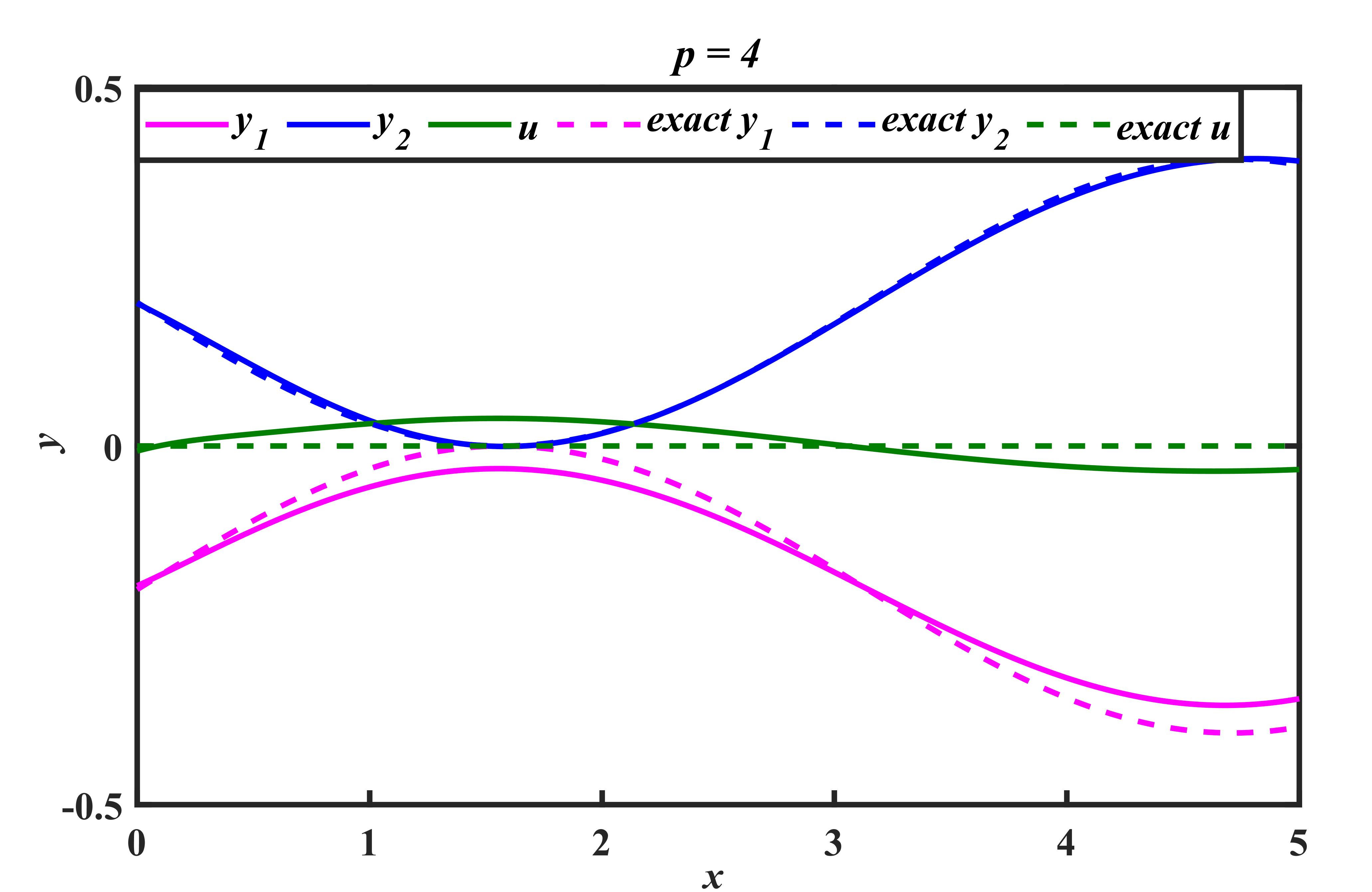}
        \caption{Supervised}
        \label{fig:ex_em}
    \end{subfigure}
    \caption{Comparison of supervised and unsupervised training.}
    \label{fig:comparison}
\end{figure}

The algebraic constraint $y_{1}^{2}-y_{2}^{2}=0$ induces two branches, $y_{1}=y_{2}$ and $y_{1}=-y_{2}$. Without supervised data, PINNs cannot distinguish between these branches and may converge to an incorrect solution manifold, as shown in Fig.~\ref{fig:comparison}(a). This highlights that the accuracy of PINN-based optimization critically depends on the availability of reliable solution data.

\subsection{Structural Analysis}

Parameter variations may lead to singular Jacobian matrices in DAE systems, causing degeneration and loss of constraint information. In such cases, numerical solvers fail to provide accurate solutions, resulting in unreliable training data for PINNs.

To address this issue, Yang~\cite{Yang2024} proposed an embedding-based structural analysis method, which transforms a degenerate system into an equivalent non-degenerate one.

\begin{theorem}[~\cite{Yang2024}]
Let $\bm{F}$ be a DAE system with rank-deficient Jacobian. Then there exists an embedded system $\bm{G}$ such that
\begin{equation}
Z_{\mathbb{R}}(\bm{F}) = \pi Z_{\mathbb{R}}(\bm{G}),
\end{equation}
where $\pi$ is a projection operator. Moreover, the differentiation index satisfies $\delta(\bm{G}) \leq \delta(\bm{F}) - (n-r)$.
\end{theorem}

This result ensures that the embedded system preserves the solution set while improving structural properties. By iteratively applying the embedding procedure until full rank is achieved, one can obtain a non-degenerate system, enabling accurate solution computation and reliable dataset construction.

\subsection{Global Error Estimation}

Although PINNs provide flexible function approximation, their solutions inevitably involve approximation errors. To quantify such errors, we consider global error estimation.

After structural analysis, a non-degenerated DAE can be transformed into an explicit ODE
\begin{equation}
\bm{x}' = \bm{f}(t,\bm{x}(t)).
\end{equation}
Applying local linearization yields
\begin{equation}
\bm{x}' = \bm{A}(t)\bm{x}(t) + \bm{b}(t),
\end{equation}
where $\bm{A}(t)$ is the Jacobian of $\bm{f}$. Let $\bm{A}(t)=\bm{A}_0+\bm{R}(t)$, where $\bm{R}(t)$ is a bounded perturbation.

\begin{theorem}[~\cite{Yang2021}]
Suppose $\bm{A}_0$ is diagonalizable with distinct eigenvalues. Then the global error satisfies
\begin{equation}
\|\Delta \bm{x}\|_{\infty} \leq \|{\bm P}\|_{\infty}\cdot \frac{\delta_{max}}{a_{1}+n r_{max}} \left(e^{(a_{1}+n r_{max})t}-1\right),
\end{equation}
where $r_{max}$ and $\delta_{max}$ denote bounds of perturbation and residual, respectively.
\end{theorem}

This bound characterizes the worst-case approximation error over the entire solution domain, providing a theoretical guarantee for evaluating the accuracy of PINN-based solutions.

\section{A Dual Physics-Informed Neural Network Architecture}
\label{sec:architecture}

For multi-task optimization problems with fixed constraints and varying objectives, the proposed architecture is designed around three requirements: the constraint solution should be computed only once and reused across different objectives; the approximation should admit a reliable error bound; and online optimization should be sufficiently fast for repeated use. Standard PINN formulations are not well suited to this setting, because the objective is embedded directly into the training process. Once the objective changes, the whole model must be retrained. To address this limitation, we decouple constraint approximation from objective optimization through a dual-network structure.

\begin{figure}[!ht]
\centering
\includegraphics[width=12cm]{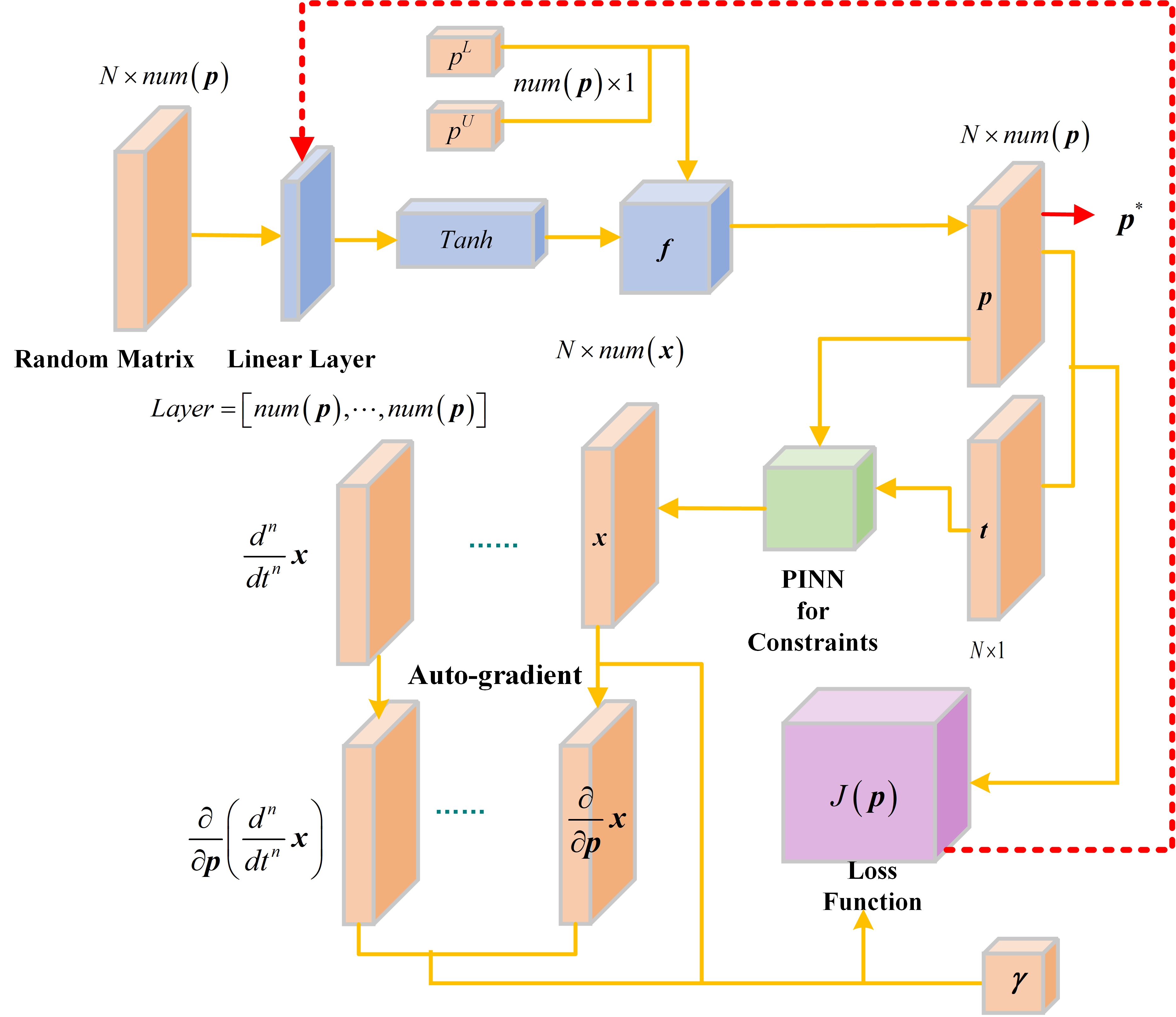}
\caption{Dual physics-informed neural network architecture for multi-task optimization}
\label{fig:pinn_model}
\end{figure}

Figure~\ref{fig:pinn_model} illustrates the proposed framework. The constraint network $\mathcal{NN}_{\text{cnstr}}(t,\bm{p};\bm{\theta}_c)$ is trained offline to approximate the parametric DAE solution mapping $(t,\bm{p}) \mapsto \bm{x}(t,\bm{p})$ over the domain $[t_0,t_f]\times[\bm{p}^L,\bm{p}^U]$. Once trained, this network is frozen and reused for all subsequent optimization tasks. For a given objective function $J(\bm{p})$, the objective network $\mathcal{NN}_{\text{obj}}(\bm{z};\bm{\theta}_o)$ is trained online to generate candidate parameters $\bm{p}$ from random seeds $\bm{z}\sim\mathcal{U}[-1,1]^d$. The resulting parameters are passed to the frozen constraint network to evaluate the corresponding state trajectory, after which the objective value is computed by automatic differentiation and numerical integration when needed.

Let the trained constraint network satisfy
\begin{equation}
\|\mathcal{NN}_{\text{cnstr}}(t,\bm{p};\bm{\theta}_c)-\bm{x}(t,\bm{p})\|_{\infty}\leq \bm{\gamma},
\quad \forall (t,\bm{p})\in [t_0,t_f]\times[\bm{p}^L,\bm{p}^U],
\end{equation}
where $\bm{\gamma}$ is the relaxation vector defined by the approximation error bound in Theorem~\ref{thm:pinn_error}. The objective network is constructed as
\begin{equation}
\bm{p}=\mathcal{NN}_{\text{obj}}(\bm{z};\bm{\theta}_o)
=\frac{1}{2}\left[\tanh(\bm{W}\bm{z}+\bm{b})\odot(\bm{p}^U-\bm{p}^L)+(\bm{p}^U+\bm{p}^L)\right],
\end{equation}
which guarantees that the generated parameters remain within the admissible bounds. Using the frozen constraint network, the approximate state trajectory is written as
\begin{equation}
\hat{\bm{x}}(t,\bm{p})=\mathcal{NN}_{\text{cnstr}}(t,\bm{p};\bm{\theta}_c)+\bm{\gamma}\odot\bm{\xi},
\qquad \bm{\xi}\sim\mathcal{U}[-1,1]^n,
\end{equation}
and the corresponding objective value is evaluated by
\begin{equation}
\hat{J}(\bm{p})
=\int_{t_0}^{t_f}\bm{\lambda}(s,\bm{p},\hat{\bm{x}}(s,\bm{p}))\,ds
+\bm{\varphi}(\bm{p},\hat{\bm{x}}(t_f,\bm{p})).
\end{equation}
The online training loss is then simply
\begin{equation}
L(\bm{\theta}_o)=\hat{J}\!\left(\mathcal{NN}_{\text{obj}}(\bm{z};\bm{\theta}_o)\right).
\end{equation}
In this way, optimization over a new objective reduces to learning a parameter-generating mapping, while the DAE constraint is handled by the offline-trained surrogate.

The equivalence between the surrogate problem and the original DAE-constrained optimization problem follows directly from the approximation accuracy of the constraint network. Define the PINN-based optimization problem as
\begin{equation}
\begin{aligned}
\min_{\bm{p}}\quad & J(\bm{p})
=\int_{t_0}^{t_f}\bm{\lambda}\bigl(s,\bm{p},\bm{x}(s,\bm{p})\bigr)\,ds
+\bm{\varphi}\bigl(\bm{p},\bm{x}(t_f,\bm{p})\bigr),\\
\text{s.t.}\quad &
\mathcal{NN}_{\text{cnstr}}(t,\bm{p};\bm{\theta}_c)-\bm{\gamma}
\leq \bm{x}(t,\bm{p})
\leq \mathcal{NN}_{\text{cnstr}}(t,\bm{p};\bm{\theta}_c)+\bm{\gamma},\\
& t\in[t_0,t_f],\quad \bm{p}\in[\bm{p}^L,\bm{p}^U].
\end{aligned}
\end{equation}
When the approximation error of $\mathcal{NN}_{\text{cnstr}}$ becomes sufficiently small, the relaxation bound $\bm{\gamma}$ converges to zero, and the surrogate optimization problem becomes equivalent to the original parametric DAE optimization problem. In practice, the exact bound is difficult to compute. We therefore use the conservative estimate
\begin{equation}
\bm{\gamma}=\max\bigl\{ |E(\zeta)-3D(\zeta)|,\; |E(\zeta)+3D(\zeta)| \bigr\},
\end{equation}
which provides a practical confidence interval for optimization. This choice is sufficiently reliable in our experiments and preserves the robustness of the dual-network formulation.

The complete training process is straightforward. The constraint network is trained once offline and then fixed. For each new objective, the objective network is initialized randomly and optimized online through repeated forward evaluation of the frozen constraint network and gradient-based updates obtained by automatic differentiation. Because no DAE solving is required during online optimization, the computational cost is significantly reduced. When higher numerical accuracy is needed, the approximate solution can be further refined by local optimization methods such as Newton iteration or, when the landscape is highly non-convex, by a randomized local search strategy.

\section{Genetic Algorithm-Based Training of the Constraint Network}
\label{sec:training}

The accuracy of the constraint network $\mathcal{NN}_{\text{cnstr}}(t,\bm{p};\bm{\theta}_c)$ is critical to the reliability of the dual PINN framework. For parametric DAE systems, constructing a representative training dataset over the entire parameter space $[\bm{p}^L,\bm{p}^U]$ is challenging. Uniform random sampling is inefficient, as it does not adapt to regions with large approximation errors and may lead to incorrect solution branches in nonlinear systems.

To address this issue, we adopt a genetic algorithm (GA)-based adaptive sampling strategy that iteratively refines the exact solution dataset $N_B$. The key idea is to identify parameter samples with large prediction errors and generate new samples in these regions using evolutionary operators. After the network reaches a prescribed error tolerance, we further apply a local correction step (e.g., Newton iteration or random walk) to obtain a high-precision solution.

\subsection{Adaptive Training Strategy with Refinement}

Let $\widetilde{\bm{p}}_i$ denote a population of parameter samples. For each sample, the prediction error is defined as
\begin{equation}
e_i = \|\mathcal{NN}_{\text{cnstr}}(t,\widetilde{\bm{p}}_i;\bm{\theta}_c) - \bm{x}^*(t,\widetilde{\bm{p}}_i)\|_{\infty},
\end{equation}
where $\bm{x}^*$ is obtained from a high-precision DAE solver. Samples with $e_i > \alpha$ are selected with probability proportional to their error, and new candidates are generated via crossover and mutation:
\begin{equation}
\widetilde{\bm{p}}_{\text{new}} = \beta \widetilde{\bm{p}}_{p1} + (1-\beta)\widetilde{\bm{p}}_{p2} + \bm{\epsilon}, 
\quad \bm{\epsilon} \sim \mathcal{N}(0,\sigma^2),
\end{equation}
with projection onto $[\bm{p}^L,\bm{p}^U]$. The corresponding exact solutions are then computed and added to $N_B$, and the network is retrained. This process repeats until the maximum prediction error falls below the tolerance $\alpha$.

Once the constraint network is trained, we can use it to generate approximate solutions for any parameter value. However, the network output may still contain small approximation errors. To obtain a final solution with higher accuracy, we apply a local correction step. For smooth objective functions, Newton iteration is efficient:
\begin{equation}
\bm{p}_{k+1}=\bm{p}_{k} - \left(\frac{\partial J(\bm{p})}{\partial \bm{p}}\bigg|_{\bm{p}=\bm{p}_{k}}\right)\Bigg/\left(\frac{\partial^{2} J(\bm{p})}{\partial \bm{p}^{2}}\bigg|_{\bm{p}=\bm{p}_{k}}\right),
\end{equation}
where the derivatives are computed using automatic differentiation of the PINN outputs. When the objective landscape is highly nonconvex or the gradient is unreliable (e.g., near multiple minima), a random walk strategy can be used instead. In our experiments, we employ Newton iteration when the solution is already close to a local optimum, and a random walk when the objective has oscillations.

\subsection{Illustrative Example}
\begin{example}\label{ex:song2022}
We consider the nonlinear ODE system from \cite{Song2022}:
\begin{equation}
\begin{array}{cl}
    obj & \underset{p}{{\min }}\,J(p) = \min\limits_{p \in [-1.2, -0.2]} -3\cdot x(t_f,p)^{3} + (1 + p)\cdot x(t_f,p) \\
    s.t. & \left\{\begin{array}{rcl}
                  \dot{x}(t,p) &=& x(t,p)^{4} - 3x(t,p)^{2} - x(t,p) + 0.4,\\
                  x(0,p) &=& p - p^{3}/3,\\ 
                  t & \in & [0, 0.9].
                 \end{array}\right.
  \end{array}
\end{equation}
This problem has a single global minimum. We first train the constraint network $\mathcal{NN}_{\text{cnstr}}$ (which approximates $x(t,p)$) using the GA-based method with $\alpha=0.001$, $M=50$ initial samples, and adding 1000 new exact solution points per generation. Figure~\ref{fig:ex_song_training} shows the training loss and maximum prediction error over 12 generations; both decrease steadily and meet the tolerance.

\begin{figure}[!ht]
\centering
\begin{subfigure}[b]{0.48\textwidth}
\includegraphics[width=\textwidth]{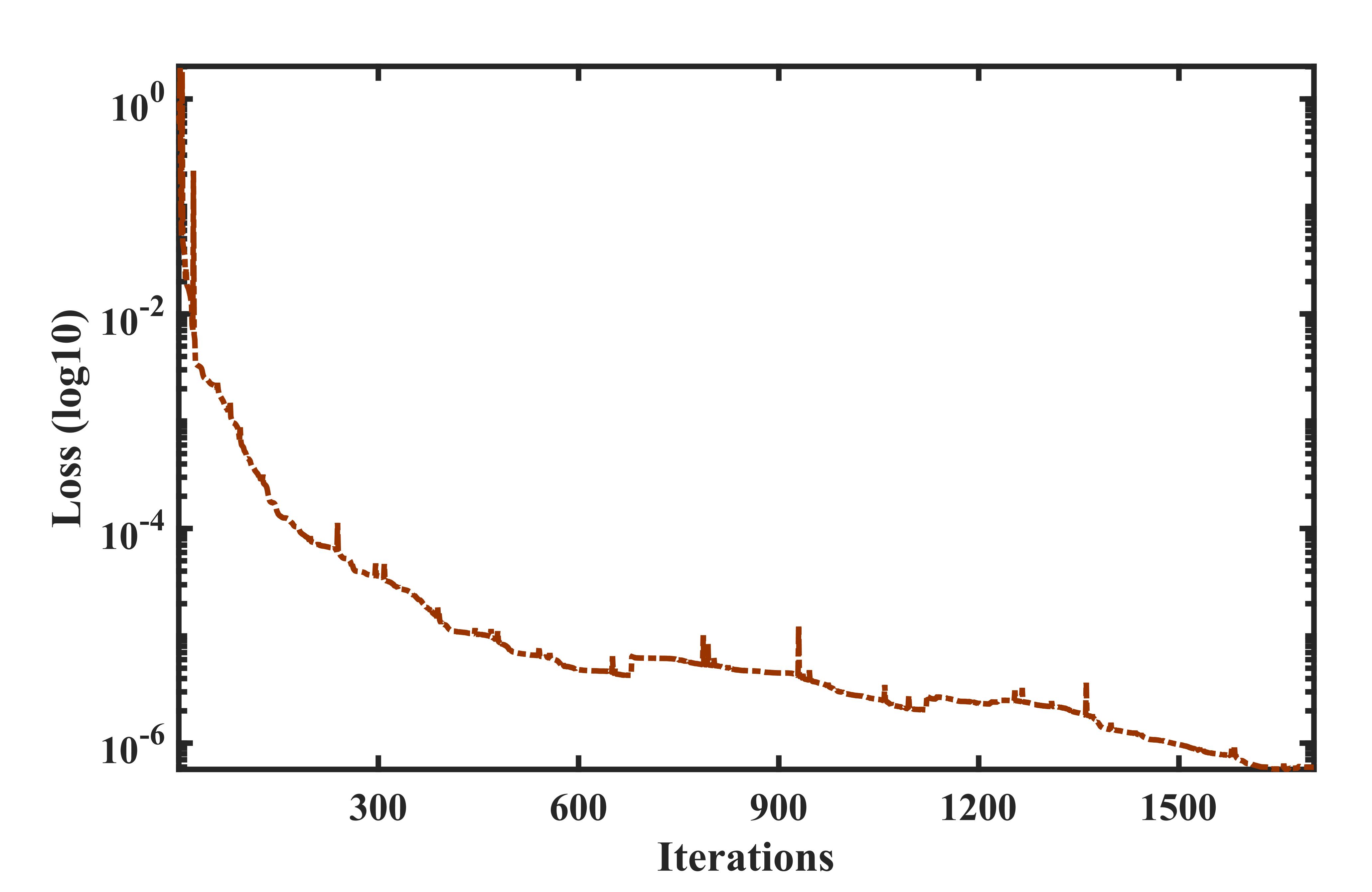}
\caption{Training loss}
\end{subfigure}
\hfill
\begin{subfigure}[b]{0.48\textwidth}
\includegraphics[width=\textwidth]{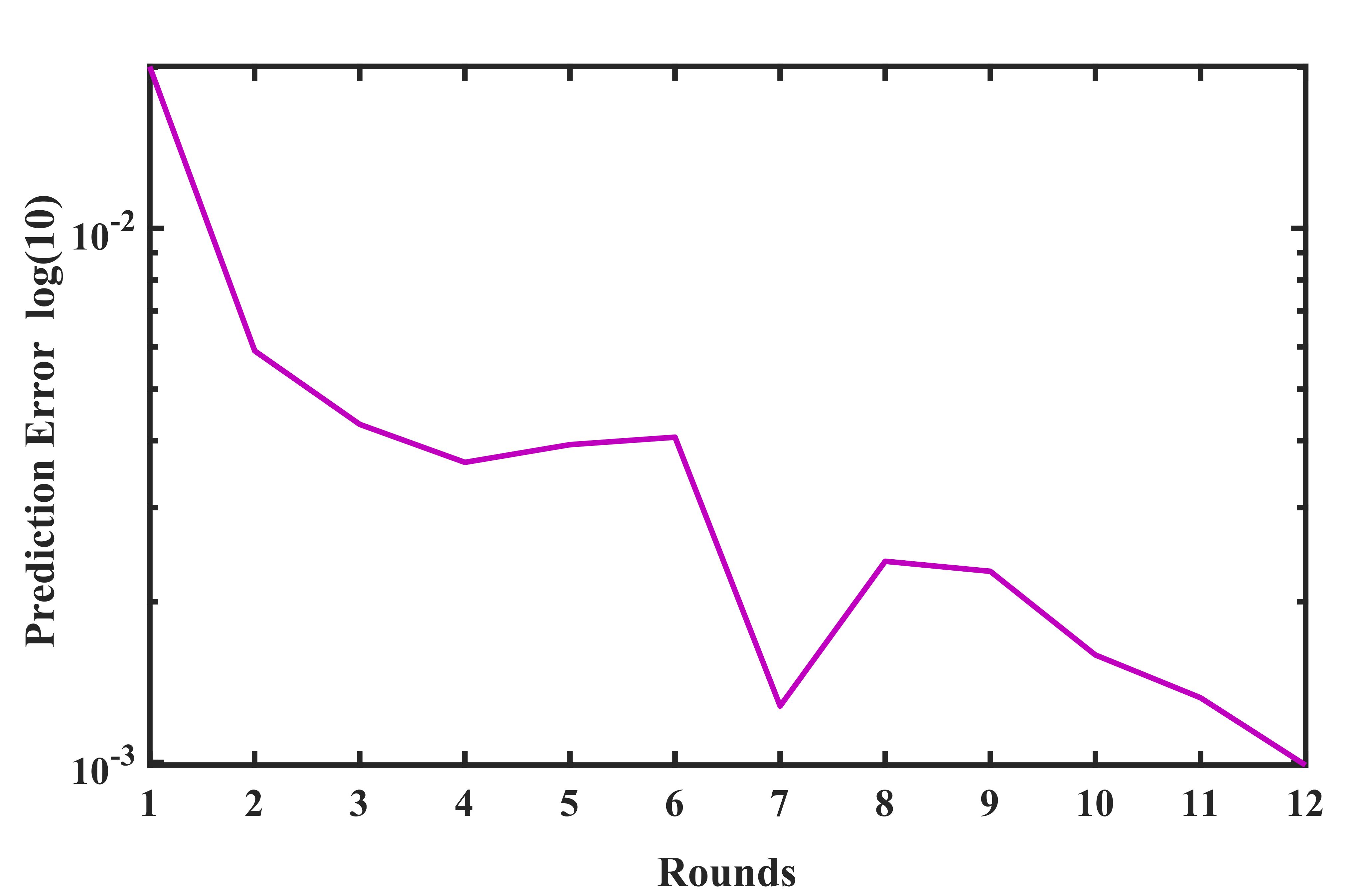}
\caption{Maximum error}
\end{subfigure}
\caption{Training progression for Example~\ref{ex:song2022}.}
\label{fig:ex_song_training}
\end{figure}

After training, the constraint network accurately predicts $x(t_f,p)$ (see Fig.~\ref{fig:ex_song_true}), enabling us to use the objective network to find approximate minima. The objective network was trained with $N=100$ random seeds and converged after only 6 iterations, yielding an approximate solution $p \approx -0.5706$ and objective value $-0.0606865$. Starting from this approximation, we performed 10 Newton iterations, which converged to $p^* = -0.570610626948189$ and $J_{\min}=-0.060686510583486$, matching the global optimum with 12-digit accuracy.

\begin{figure}[!ht]
\centering
\begin{subfigure}[b]{0.48\textwidth}
\includegraphics[width=\textwidth]{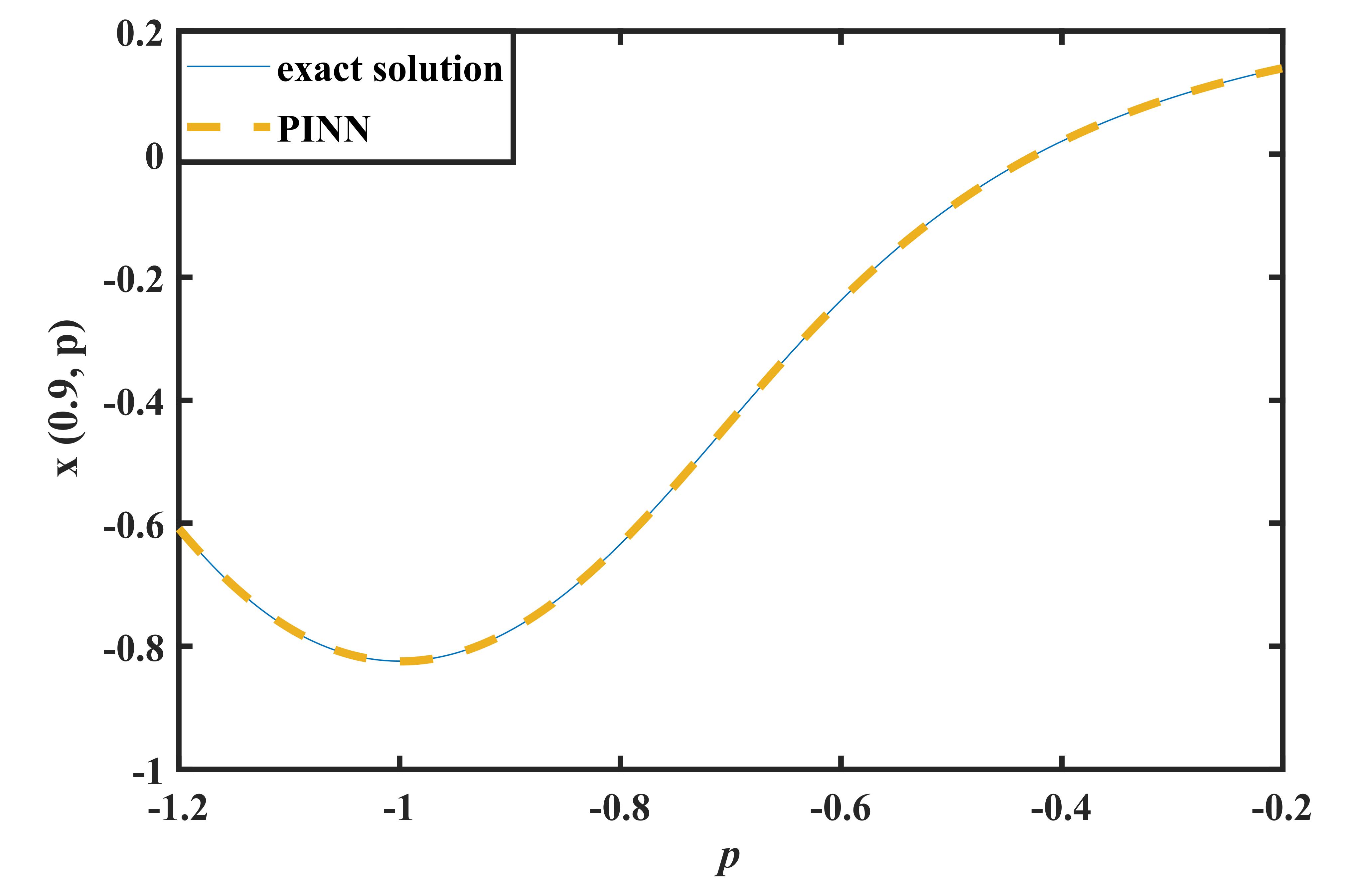}
\caption{$x(t_f,p)$: exact vs. predicted}
\end{subfigure}
\hfill
\begin{subfigure}[b]{0.48\textwidth}
\includegraphics[width=\textwidth]{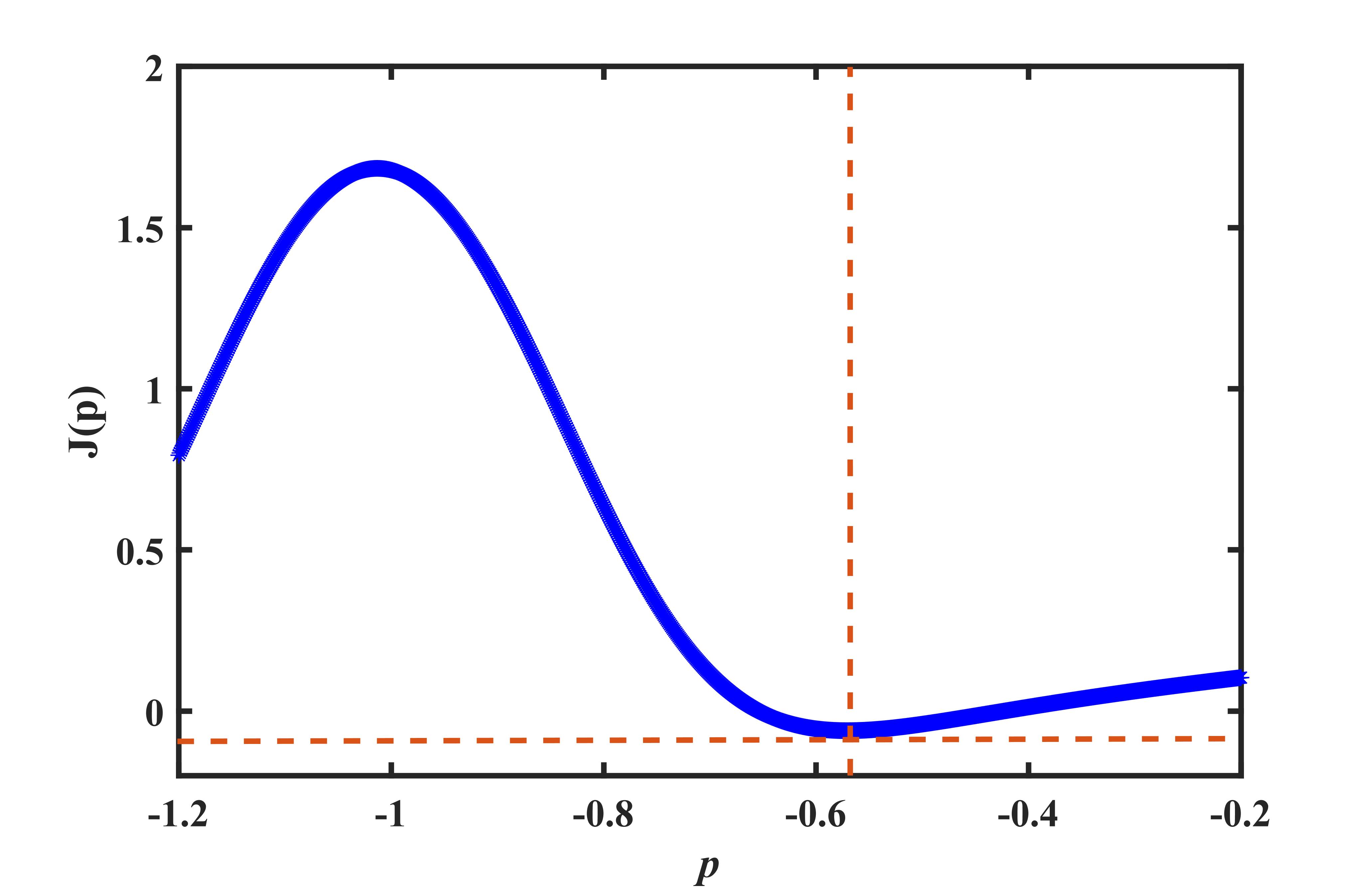}
\caption{Objective function $J(p)$}
\end{subfigure}
\caption{Exact and predicted solutions for Example~\ref{ex:song2022}.}
\label{fig:ex_song_true}
\end{figure}

A comparison with existing methods is given in Table~\ref{tab:song_efficiency}. Our approach achieves the same optimum as the relaxation methods but with far fewer iterations, demonstrating the efficiency of the combined GA training and local refinement.

\begin{table}[!ht]
\centering
\caption{Comparison of solving efficiency for Example~\ref{ex:song2022}.}
\label{tab:song_efficiency}
\begin{tabular}{c|c|c}
\hline
Method & Minimum Objective Value & Iteration Count \\
\hline
New OB relaxation~\cite{Song2022} & $-0.0607$ & 23 \\
SBM relaxation~\cite{Scott2012} & $-0.0607$ & 33 \\
Our method & $-0.0607$ & \textcolor{red}{6} \\
\hline
\end{tabular}
\end{table}
\end{example}

\subsection{Error Bound}

\begin{theorem}\label{thm:pinn_error}
For sufficiently dense training samples, the prediction error satisfies
\begin{equation}
\|\Delta \bm{x}\|_{\infty}
\le
\|\bm{P}\|_{\infty}
\cdot
\frac{\max\{|E(\zeta)-3D(\zeta)|,\;|E(\zeta)+3D(\zeta)|\}}{\bar{a}_1 + n r_{\max}}
\left[e^{(\bar{a}_1 + n r_{\max})\hbar} - 1\right],
\end{equation}
where $\zeta$ denotes residual samples, $\bar{a}_1$ is the maximum real part of the linearized system's eigenvalues, $r_{\max}$ bounds the perturbation, and $\hbar$ is the maximum time step between adjacent sample points.
\end{theorem}

This result provides a theoretical justification for the relaxation bound
\begin{equation}
\bm{\gamma}=\max\{|E(\zeta)-3D(\zeta)|,\;|E(\zeta)+3D(\zeta)|\},
\end{equation}
which we use in the dual-network optimization framework. The bound guarantees that the PINN solution lies within a confidence interval around the true solution with high probability, ensuring the reliability of the subsequent optimization step.

\section{Experiments}
\label{sec:experiments}

This section evaluates the proposed dual-network optimization framework on representative parametric DAE systems. The experiments focus on two aspects: (i) approximation accuracy of the constraint network, and (ii) efficiency and reliability of the resulting optimization process.

\subsection{Experimental Setup}

All experiments are implemented in Python on a workstation with an Intel(R) Core(TM) i7-14700KF CPU (3.40 GHz) and 32 GB DDR4 RAM. The constraint network $\mathcal{NN}_{\text{cnstr}}$ is constructed using a fully connected neural network with 5--7 hidden layers and 128 neurons per layer. The training dataset consists of three components: initial condition data $N_I$, collocation points $N_F$, and exact solution data $N_B$ generated by a high-precision DAE solver.

The GA-based training method described in Section~\ref{sec:training} is used to adaptively refine $N_B$, with error tolerance typically set to $\alpha=10^{-3}$. All variables and parameters are normalized to improve numerical stability.

\subsection{Case Study: Biochemical Reaction System}

We consider a biochemical reaction network modeling potassium uptake regulation in \textit{E. coli}~\cite{Scott2015}. The system is described by a parametric DAE with both differential and algebraic equations, as given in Equation~(\ref{opt_goal}).

The objective is to minimize the discrepancy between model predictions and experimental measurements:
\begin{equation}
J(\bm{p})=
\frac{1}{7}\sum_{\ell}
\left[
\left(\frac{x_1(t_\ell,\bm{p})-\hat{x}_1}{\hat{x}_1}\right)^2
+
\left(\frac{x_2(t_\ell,\bm{p})-\hat{x}_2}{\hat{x}_2}\right)^2
\right],
\end{equation}
where $\bm{p}=(k_1,k_3)$ are the unknown parameters to be estimated.

\subsubsection{Constraint Network Training}
\begin{figure}[!ht]
\centering
\includegraphics[width=12cm]{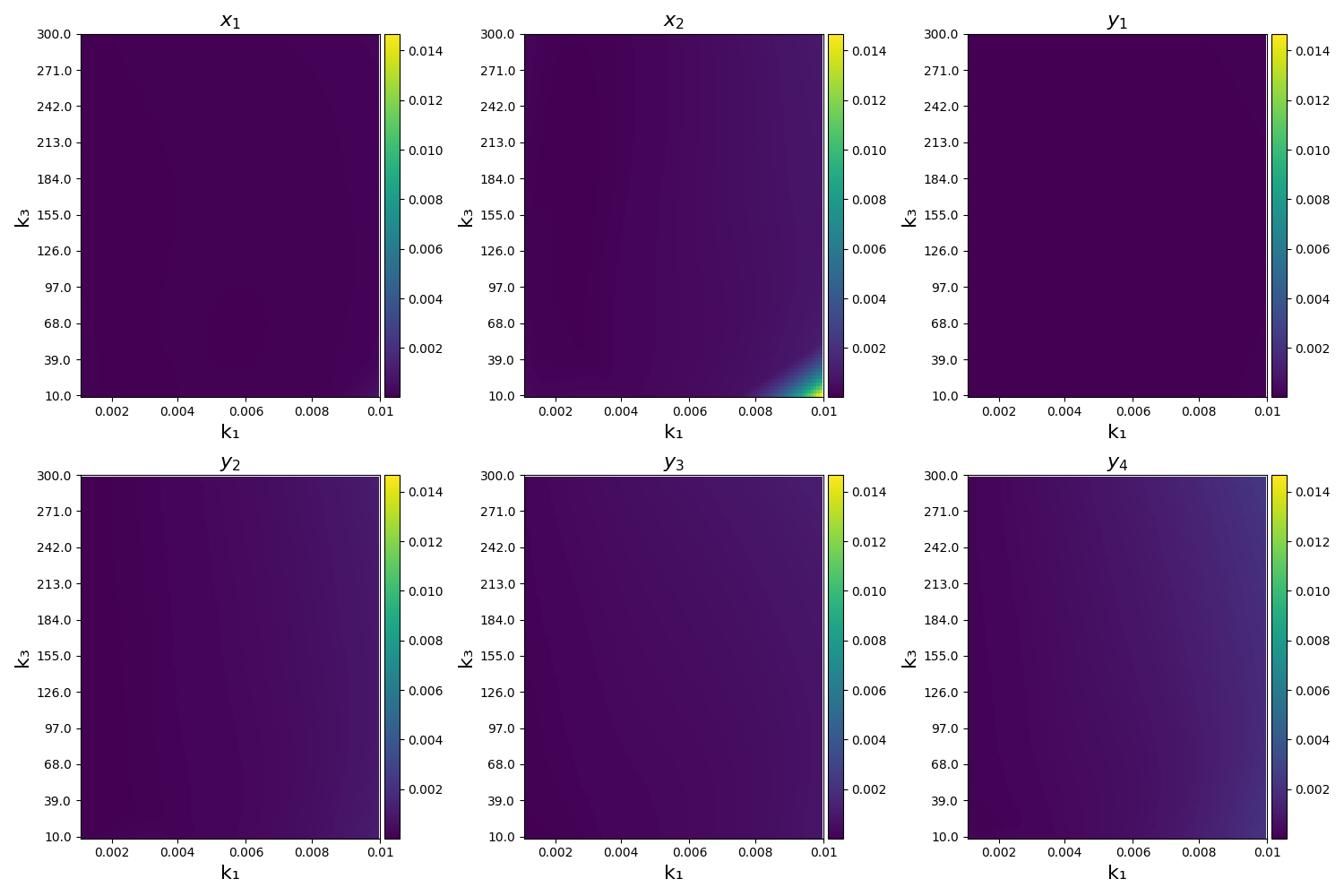}
\caption{Maximum absolute error of the constraint network for the \textit{E. coli} model.}
\label{fig:net_acc}
\end{figure}

The constraint network was trained offline using the GA-based method. Figure~\ref{fig:net_acc} shows the maximum absolute error distribution of the trained network across the parameter space. The results indicate high accuracy except for a small region around $(k_1=0.01,k_3=10)$, which can be further improved by adding more samples in that area.

\subsubsection{Optimization Results}
\begin{figure}[!ht]
\centering
\includegraphics[width=8cm]{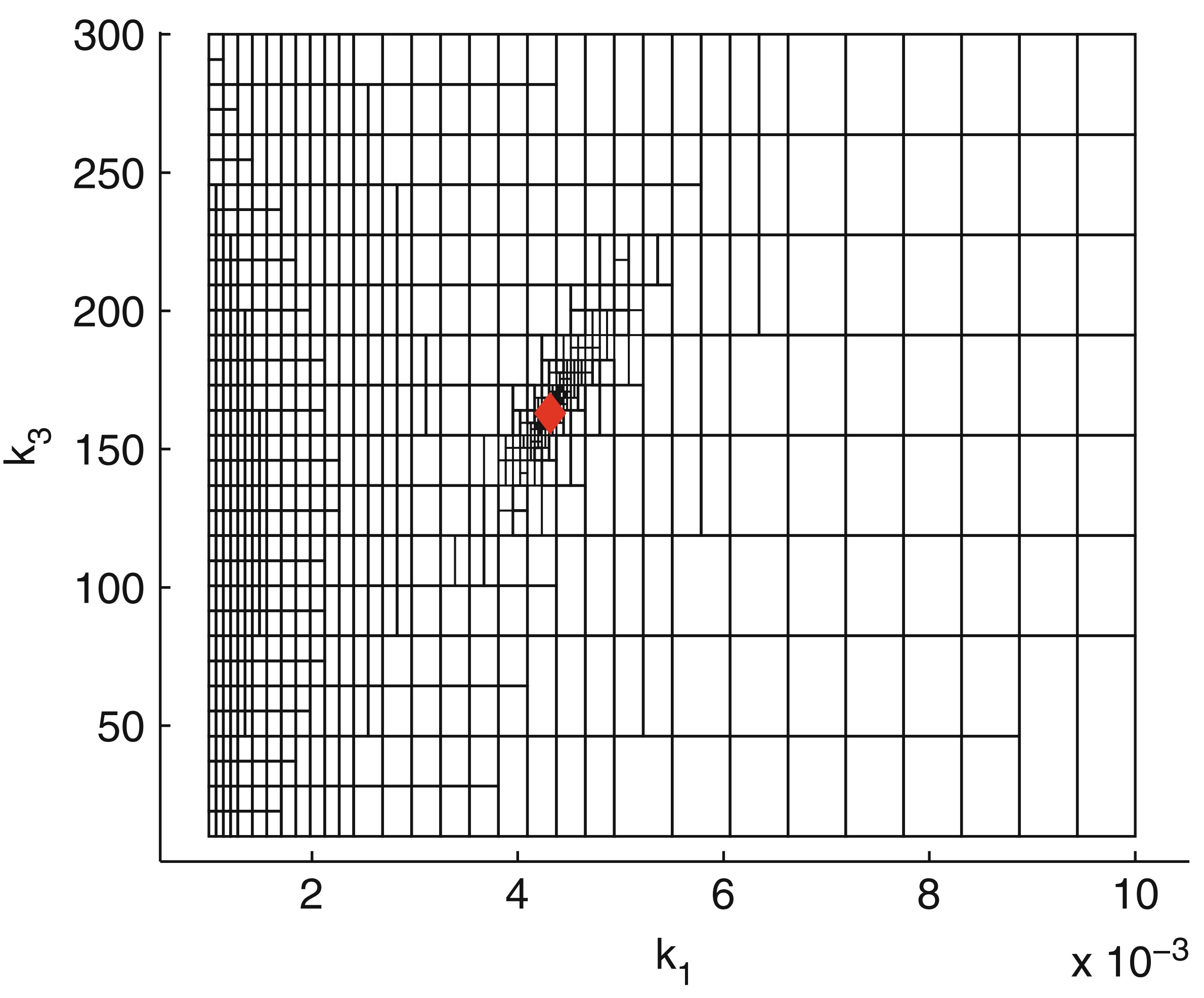}
\caption{Traversal process of the $\epsilon$-global method in~\cite{Scott2015}.}
\label{fig:eG}
\end{figure}

\begin{figure}[!ht]
\centering
\includegraphics[width=10cm]{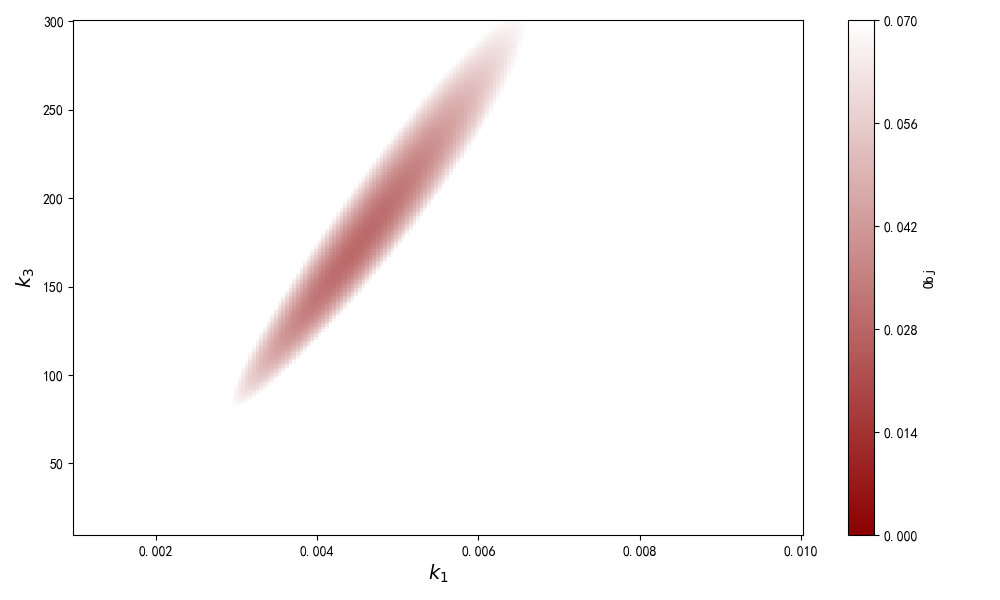}
\caption{Objective function value distribution predicted by the dual network architecture.}
\label{fig:dis}
\end{figure}

After training, the objective network was trained online to find the optimal parameters. The entire online process, including training and prediction, took only 0.945 seconds. Figure~\ref{fig:eG} illustrates the traversal process of the $\epsilon$-global method from the literature, while Figure~\ref{fig:dis} shows the distribution of objective function values predicted by our dual-network architecture. The predicted region of optimal solutions aligns closely with the results reported in~\cite{Scott2015}, confirming the accuracy of our method.

A quantitative comparison is provided in Table~\ref{table:result_coil}. The solution obtained by our method is close to the global optimum found by the $\epsilon$-global method and significantly better than the local solution obtained by the method in~\cite{Kremling2004}. A final correction step using a random walk (300 iterations, step size 10\% of the parameter range) improved the solution further, resulting in the refined optimal solution reported in the table. The total online time (training, prediction, and correction) was approximately 1 second, significantly faster than the 52 seconds required by the branch-and-bound method.

\begin{table}[!ht]
\centering
\caption{Comparison of optimization results for the \textit{E. coli} model.}
\label{table:result_coil}
\begin{tabular}{|c|c|c|}
\hline
Method & Parameters $(k_1, k_3)$ & $Obj$ \\
\hline
$\epsilon$-global method~\cite{Scott2015} & $(4.31348\times10^{-3}, 162.9297)$ & $0.029096$ \\
Method in~\cite{Kremling2004} & $(2.9\times10^{-3}, 90)$ & $0.0712$ \\
Our method (predicted) & $(4.4881\times10^{-3}, 1.7231\times10^{2})$ & $0.0296$ \\
Our method (corrected) & $(4.32191\times10^{-3}, 1.632459\times10^{2})$ & $0.029089$ \\
\hline
\end{tabular}
\end{table}

\subsection{Performance Analysis}

To evaluate scalability, we consider a high-dimensional linear DAE system from Example~\ref{ex:analysis}~\cite{Bajaj2020}. The experimental setup follows the parameters described in Section~\ref{sec:experiments}.

\begin{example}\label{ex:analysis}
The optimization problem is:
\[
\begin{array}{cc}
obj & \underset{\bm{p}}{{\min }}~~~\frac{1}{n}\cdot(\bm{x}(1)-e^{-5\cdot\bm{p}})^{2}\\
s.t. & \left\{\begin{array}{ccc}
  \dot{\bm{x}}(t) & = & \bm{A}\cdot\bm{x}(t) + \bm{I}\cdot\bm{p}\\
    \bm{x}(0) & = & \bm{1}_{n\times 1}
\end{array}\right.\\
\end{array}
\]
where $\bm{A}$ is a bidiagonal matrix with entries $-5n$ on the diagonal and $5n$ on the subdiagonal, and $\bm{p}\in\mathbb{R}^n$.
\end{example}

Table~\ref{table:result_t} summarizes the computational performance and accuracy for varying dimensions $n$. The offline training time increases with dimension, but the online prediction and correction times remain low and scale approximately linearly. In contrast, the branch-and-bound method's runtime grows exponentially and becomes impractical for $n>6$. The solution accuracy remains high across all tested dimensions, with prediction errors on the order of $10^{-7}$.

\begin{table}
\caption{Performance Comparison for Example\label{table:result_t}}
\centering
\begin{tabular}{|c|c|c|c|p{40pt}|c|c|p{51pt}|p{51pt}|}
\hline
$n$ &  \begin{tabular}{@{}c@{}}Training\\Cost (s)\end{tabular} & \begin{tabular}{@{}c@{}}Prediction\\Cost (s)\end{tabular} & \begin{tabular}{@{}c@{}}Correction\\Cost (s)\end{tabular} & \begin{tabular}{@{}c@{}}B$\&$B\\Cost (s)~\cite{Bajaj2020}\end{tabular} & Predicted $J(\bm{p})$ & Predicted $\bm{p}$ & Optimal $\bm{p}^{*}$ \\
\hline
$2$ & $96.1$ & $0.6$ & $0.003$ & $1.4$ & $6.50e-08$ & $\begin{array}{c} [0.5714, \\ 0.4541] \end{array}$ & $\begin{array}{c} [0.5721,\\0.4544\end{array}$\\
\hline
$3$ & $435.2$ & $1.2$ &$0.003$ & $8.1$ & $1.17e-07$ &$\begin{array}{c}[0.6325, \\ 0.5137,\\ 0.4474] \end{array}$&$\begin{array}{c}[0.6330,\\0.5141,\\0.4481] \end{array}$\\
\hline
$4$ & $1452.3$ &$1.3$ & $0.011$ & $19.4$ & $1.28e-07$ &$\begin{array}{c}[0.6751,\\ 0.5564,\\ 0.4902,\\ 0.4437]\end{array}$&$\begin{array}{c}[0.6771,\\ 0.5572,\\0.4904,\\0.4444]\end{array}$\\
\hline
$6$ & $5991.9$ &$1.5$ & $0.005$ & $90.6$ & $8.07e-07$ & $\begin{array}{c}[0.7396,\\ 0.6170,\\ 0.5484,\\ 0.5014,\\ 0.4667,\\ 0.4392]\end{array}$& $\begin{array}{c}[0.7404,\\ 0.6187,\\ 0.5506,\\0.5038,\\0.4685,\\ 0.4403]\end{array}$\\
\hline
$8$ & $18424.8$ & $2.7$ & $0.013$ & $934.6$ & $1.60e-07$ & $\begin{array}{c}[0.7852,\\0.6633,\\0.5949,\\0.5473,\\0.5109,\\0.4823,\\0.4578,\\0.4372]\end{array}$& $\begin{array}{c}[0.7859,\\0.6636,\\0.5949,\\0.5474,\\ 0.5113,\\0.4825,\\0.4582,\\0.4377]\end{array}$\\
\hline
$10$ & $53472.4$ & $4.8$ & $0.008$ & $6083.5$ & $2.92e-07$ & $\begin{array}{c}[0.8241,\\ 0.6981,\\ 0.6285, \\0.5809, \\0.5443, \\0.5148, \\0.4902, \\0.4692, \\0.4513,\\0.4348]\end{array}$ & $\begin{array}{c}[0.8241,\\ 0.6984,\\0.6287,\\0.5811,\\ 0.5445,\\ 0.5150,\\0.4904,\\0.4694,\\0.4515,\\ 0.4350]\end{array}$\\
\hline
\end{tabular}
\end{table}

\section{Results and Discussion}

The experimental results demonstrate the following key advantages of the proposed framework:

\textbf{Efficiency}: The dual-network architecture decouples constraint approximation from objective optimization, allowing the expensive constraint network to be trained only once. Online optimization for new objectives then requires only training a lightweight objective network, resulting in a speedup of one to two orders of magnitude compared to conventional branch-and-bound methods, especially for high-dimensional problems.

\textbf{Accuracy}: The GA-based training strategy ensures that the constraint network achieves high accuracy across the entire parameter space, with prediction errors typically below $10^{-3}$. The final solution can be further refined using local optimization techniques, yielding results comparable to or better than global deterministic methods.

\textbf{Global Optimization}: The architecture does not rely on gradient information from the objective function and incorporates random perturbations during training, enabling it to escape local minima and identify multiple promising solution candidates, as shown in the biochemical example.

\textbf{Scalability}: While the offline training cost grows with problem size, the online phase scales nearly linearly, making the approach suitable for real-time and multi-task applications where constraints are fixed but objectives change frequently.

\bibliography{sn-bibliography}
\end{document}